\documentclass[conference]{IEEEtran}
\IEEEoverridecommandlockouts
\usepackage{cite}
\usepackage{amsmath,amssymb,amsfonts}
\usepackage{algorithmic}
\usepackage{graphicx}
\usepackage{textcomp}
\usepackage{xcolor}
\usepackage[raggedright]{sidecap}
\usepackage[singlelinecheck=false,justification=justified]{caption}
\usepackage{booktabs}
\usepackage{multirow}

\usepackage{caption}
\usepackage{subcaption}

\usepackage{sidecap}
\sidecaptionvpos{figure}{t}
\usepackage[colorlinks]{hyperref}

\newcommand\blfootnote[1]{%
  \begingroup
  \renewcommand\thefootnote{}\footnote{#1}%
  \addtocounter{footnote}{-1}%
  \endgroup
}

\def\BibTeX{{\rm B\kern-.05em{\sc i\kern-.025em b}\kern-.08em
    T\kern-.1667em\lower.7ex\hbox{E}\kern-.125emX}}
\begin{document}

\title{Single-branch Network for Multimodal Training}

\author{Muhammad Saad Saeed$^{1}$\textsuperscript{\textdagger}, Shah Nawaz$^{2}$\textsuperscript{\textdagger}, Muhammad Haris Khan$^{3}$, \\ Muhammad Zaigham Zaheer$^{3}$, Karthik Nandakumar$^{3}$, Muhammad Haroon Yousaf$^{4}$, Arif Mahmood$^{5}$  \\
$^{1}$Swarm Robotics Lab NCRA, 
$^{2}$Deutsches Elektronen-Synchrotron DESY, \\
$^{3}$Mohamed bin Zayed University of Artificial Intelligence,
$^{4}$University of Engineering and Technology Taxila,\\
$^{5}$Information Technology University
}


\maketitle

\begin{abstract}
\blfootnote{\textsuperscript{\textdagger}Equal contribution}
With the rapid growth of social media platforms, users are sharing billions of multimedia posts containing audio, images, and text. 
Researchers have focused on building autonomous systems capable of processing such multimedia data to solve challenging multimodal tasks including cross-modal retrieval, matching, and verification.
Existing works use separate networks to extract embeddings of each modality to bridge the gap between them.
The modular structure of their branched networks is fundamental in creating numerous multimodal applications and has become a defacto standard to handle multiple modalities. In contrast, we propose a novel single-branch network capable of learning discriminative representation of unimodal as well as multimodal tasks without changing the network. An important feature of our single-branch network is that it can be trained either using single or multiple modalities without sacrificing performance. 
We evaluated our proposed single-branch network on the challenging multimodal problem (face-voice association) for cross-modal verification and matching tasks with various loss formulations.
Experimental results demonstrate the superiority of our proposed single-branch network over the existing methods in a wide range of experiments. 
Code: \href{https://github.com/msaadsaeed/SBNet}{https://github.com/msaadsaeed/SBNet}
\end{abstract}

\begin{IEEEkeywords}
Multimodal data, Two-branch networks, Face-voice association, Cross-modal verification and matching
\end{IEEEkeywords}

\section{Introduction}
Recent years have seen a surge in multimodal data containing various modalities. Generally, users combine image, text, audio, or video data to express opinions on social media platforms. 
The combinations of these media types have been extensively studied to solve several multimodal tasks including cross-modal retrieval~\cite{wang2016learning,nawaz2021cross}, cross-modal verification~\cite{saeed2022learning,nagrani2018seeing}, multimodal named entity recognition~\cite{arshad2019aiding,moon1078}, visual question answering~\cite{anderson2018bottom,fukui2016multimodal}, image captioning~\cite{vinyals2015show,popattia2022guiding}, and  multimodal classification~\cite{gallo2017multimodal,kiela2018efficient,gallo2018image}.
In the existing multimodal systems, neural network based mappings have been commonly used to bridge the semantic gap between multiple modalities by building a joint representation.
For example, separate independent networks are leveraged to predict features of each modality and a supervision signal is employed to learn joint representation in a two-branch network~\cite{wang2016learning,faghri2018vse++,nagrani2018seeing,nagrani2018learnable,saeed2022learning,saeed2022fusion,nawaz2021cross,kim2018learning,nawaz2022semantically,nawaz2019these}. 
In addition, multimodal systems have leveraged Transformers to learn joint representation in a two-branch setting~\cite{lu2019vilbert,tan2019lxmert}.
The modular nature of the two-branch network is instrumental in developing various multimodal applications and has become a common practice to process multiple modalities.
However, embeddings extracted from modality-specific networks share many semantic similarities. For example, the gender, nationality, and age of speakers are represented by their audio and visual signatures~\cite{nagrani2018learnable,saeed2022fusion}.
Therefore a fundamental question arises: \textit{can multimodal joint representations be learned with only a single-branch network?}

\begin{figure}[t!]
\centering
\includegraphics[scale=0.6]{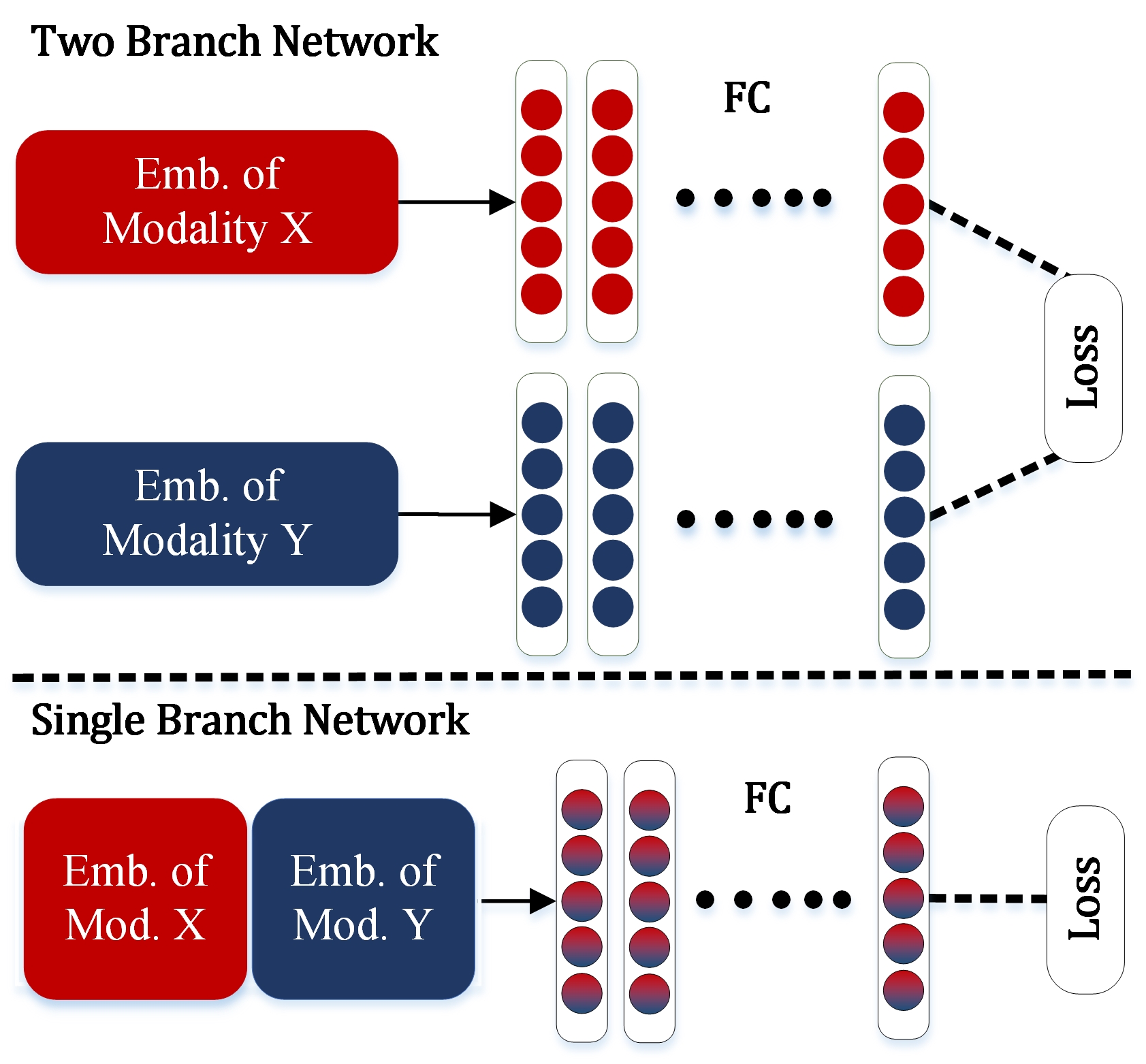}
   \caption{
   (Top) The existing two-branch networks employ independent modality-specific branches to learn a joint representation from the embeddings of modality X and Y. (Bottom) In contrast, the proposed single-branch network leverages only one branch to learn similar representations.
   }
\label{fig:motivation}
\end{figure}

\begin{figure*}[!t]
     \centering
     \begin{subfigure}[b]{0.48\textwidth}
         \centering
         \includegraphics[width=\textwidth]{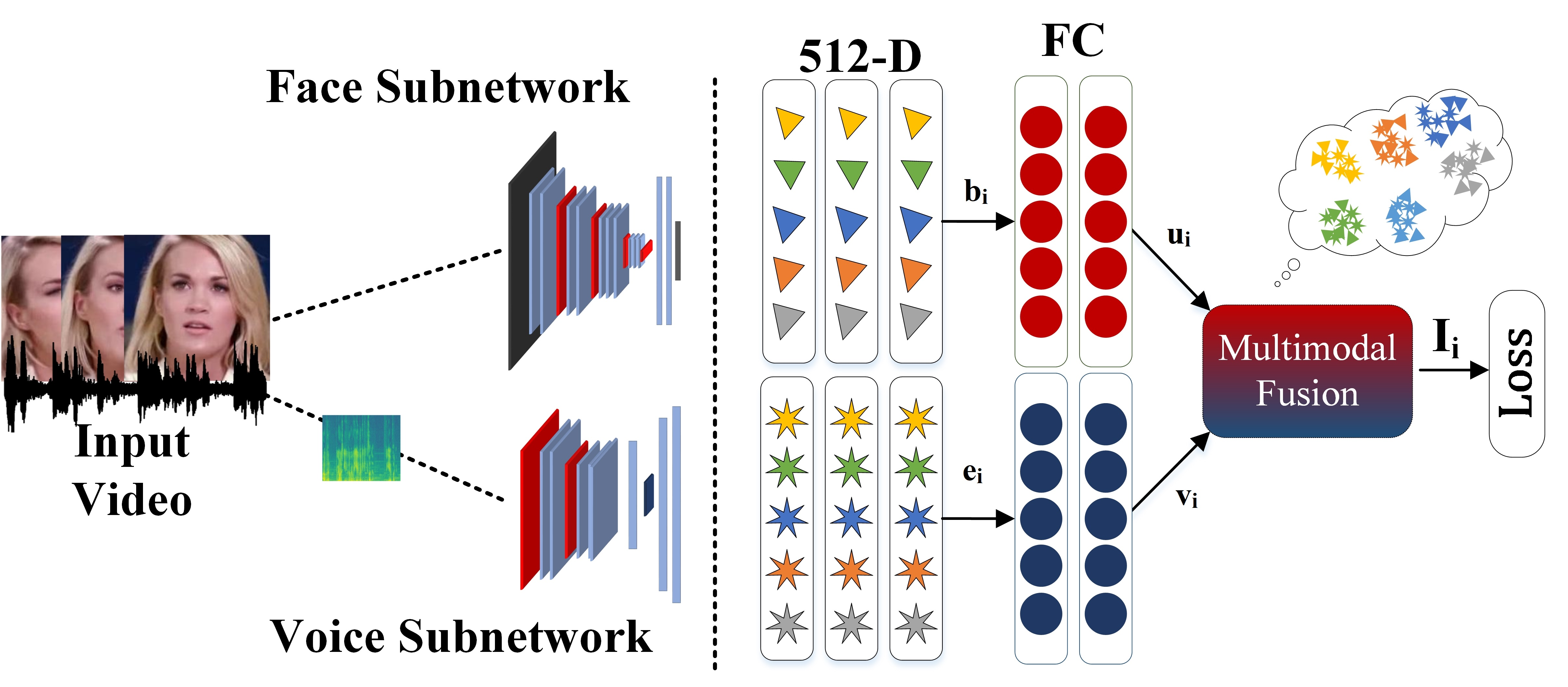}
         \caption{Conventional Two-branch Network}
         \label{fig:two-branch}
     \end{subfigure}
     \begin{subfigure}[b]{0.48\textwidth}
         \centering
         \includegraphics[width=\textwidth]{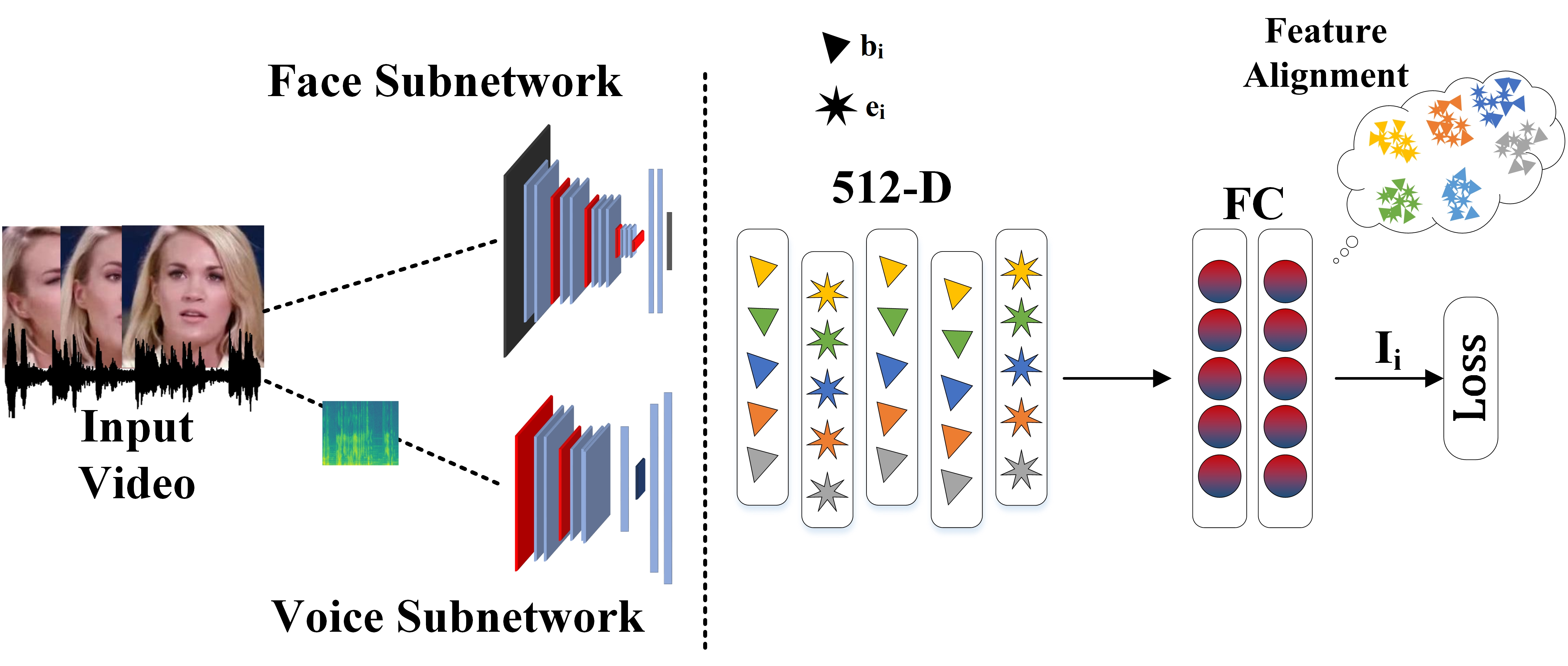}
         \caption{Our Single-branch Network}
         \label{fig:single-branch}
     \end{subfigure}
        \caption{
        (a) Two independent modality-specific embedding networks to extract features (left) and  a conventional two-branch network (right) having \textbf{two} independent modality-specific branches to learn discriminative joint representations  of the multimodal task.
        (b) Proposed  network with a \textbf{single} modality-invariant branch.
        }
        \label{fig:main_arch}
    
\end{figure*}

To investigate it, we introduce \textbf{S}ingle \textbf{B}ranch \textbf{Net}work (SBNet), a method to learn discriminative joint representation for multimodal tasks. 
The proposed SBNet consists of the following three components: 1) embedding extraction of each modality with task-specific pre-trained networks, 2) a series of modality-invariant fully connected layers in a single-branch to learn joint multimodal representations, and 3)
an extensive evaluation of proposed single-branch network after employing different loss formulations. 
By leveraging the fact that extracted embeddings share the same semantics, our formulation processes multiple modalities with a single branch to learn joint representations for multimodal tasks.
This also makes it possible to learn unimodal representations without modifying the underlying network.
Fig.~\ref{fig:motivation} compares a typical two-branch network with our proposed single-branch network. 
We use the same input features under various loss formulations for both single-branch and two-branch networks on a popular multimodal task, namely face-voice (F-V) association. 
Nagrani et al. \cite{nagrani2017voxceleb} introduced F-V association to multimodal community with the creation of a large-scale audio-visual dataset (VoxCeleb$1$~\cite{nagrani2017voxceleb}).
In addition, Nawaz et al.~\cite{nawaz2021cross} extended F-V association by analyzing the impact of language on the association.
Since then, it has gained significant research interest~\cite{horiguchi2018face,kim2018learning,nagrani2018learnable,nagrani2018seeing,nawaz2019deep,wen2021seeking,wen2018disjoint,ning2021disentangled,zheng2021adversarial,saeed2022learning,saeed2022fusion,shah2023speaker}.
All these existing methods also leverage two-branch networks to establish the correlation between faces and voices with the help of cross-modal verification and matching tasks. 
Thus, F-V association is considered a suitable benchmark application for comparison among single and two-branch networks. In addition, F-V association datasets provide unimodal tasks, for example, speaker verification which is important to showcase the capability of our single-branch network to train on unimodal or multimodal data without changing the underlying network.

We summarize our key contributions as follows: 1) We propose a single-branch network to learn multimodal discriminative joint representations. 2) We present rigorous comparisons of our single with two-branch networks under same inputs and various loss formulations on cross-modal verification and matching tasks on a large-scale VoxCeleb$1$ dataset. Experimental results show that the single-branch network outperforms two-branch network on F-V association task. Further, we note that our method performs favourably against the existing state-of-the-art methods on the F-V association task. 3) We perform a thorough ablation study to analyze the impact of different components.

\section{Overall Framework}
\label{section:overall_framework}
We develop a single-branch network to learn a joint representation for a multimodal task of F-V association (see Fig.~\ref{fig:main_arch}).


\subsection{Single-branch Network}
\label{subsection:Preliminaries}
Our aim is to learn joint representation to perform a multimodal task that establishes an association between the faces and voices of different speakers in cross-modal verification and matching.
Given that we have $N$ occurrences of face-voice pairs, $\mathcal{D}=\{(x_{i}^{f},x_{i}^{v})\}_{i=1}^{N}$, where $x_{i}^{f}$ and $x_{i}^{v}$ are the face and voice samples of the $i_{th}$ occurrence, respectively. 
Each face-voice pair $(x_{i}^{f},x_{i}^{v})$  has a label $y_{i}\in\{0,1\}$, where $y_{i}=1$ if a pair belongs to the same speaker and $y_{i}=0$ if it belongs to different speakers. 
Cross-modal learning aims at mapping both faces and voices into a common but discriminative embedding space, where they are adequately aligned and occurrences from the same identity are nearby while those from a different identity are far apart.
Previous works approach the problem by assuming that modalities have different representations and structures, and have therefore leveraged independent modality-specific branches to learn discriminative joint representation~\cite{saeed2022fusion,saeed2022learning,nagrani2018learnable,kim2018learning,nagrani2018seeing,nawaz2021cross}. 
We, on the other hand, approach the problem by assuming shared semantics, such as gender, nationality and age for each modality. 
To this end, the face and voice embeddings are extracted from modality-specific networks. 
Precisely, face embeddings ($\mathbf{b}_i \in \mathbb{R}^{F}$) and voice embeddings ($\mathbf{e}_i \in \mathbb{R}^{V}$) are extracted from the penultimate layers of pre-trained CNNs~\cite{schroff2015facenet,xie2019utterance}.
Afterwards, face ($\mathbf{b}_i$) and voice embeddings ($\mathbf{e}_i$) are projected with a single modality-invariant network consisting of two fully-connected layers with ReLU activation function to a new $d$-dimensional embedding space $\mathbf{I}_i$.
In addition, batch normalization~\cite{ioffe2015batch} is applied right after the last linear layer. 
Using this configuration, it becomes unnecessary to create a separate branch for each modality because embeddings taken from pre-trained CNNs share the same semantics.
In other words, our SBNet learns representation irrespective of the input modality.  
Moreover, the proposed configuration is useful for learning unimodal representation with the same two fully connected layers for either audio or face recognition tasks.

\subsection{Two-branch Network.} 
In order to fairly compare our SBNet, we also use a well known two-branch network~\cite{saeed2022fusion,saeed2022learning,nagrani2018learnable} for comparison purposes.
In this configuration, face and voice embeddings are first extracted from pre-trained CNNs, denoted as $\mathbf{b}_i$ and $\mathbf{e}_{i}$ respectively.
Afterwards, face and voice embeddings are input to two independent branches, with each modality specific branch $\mathbf{u}_i$ and $\mathbf{v}_i$ respectively. 
Both $\mathbf{u}_i$ and $\mathbf{v}_i$  are then L$2$ normalized and later fused, e.g., using an attention mechanism \cite{saeed2022fusion,saeed2022learning}, to obtain $\mathbf{l}_{i}$.




\subsection{Loss Formulations}
In this section, we formally overview several existing loss formulations typically employed in existing face-voice association methods.

\noindent \textbf{Fusion and Orthogonal Projection.} It imposes orthogonality constraints on the fused embeddings to explicitly minimize intra-identity variation while maximizing inter-identity separability~\cite{ranasinghe2021orthogonal,saeed2022fusion,saeed2022learning}. These constraints complement better with the innate angular characteristic of cross entropy (CE) loss. 
The loss is formulated as follow:
\begin{equation}
  \mathcal{L}_{OC}  =  1 - \sum_{i,j \in B, y_{i}=y_{j}} \langle \mathbf{l}_{i},\mathbf{l}_{j}\rangle + \displaystyle\left\lvert \sum_{i,j \in B, y_{i} \neq y_{k}} \langle \mathbf{l}_{i},\mathbf{l}_{k}\rangle\right\rvert,
    \label{Eq:OC}
\end{equation}

\noindent where $ \langle.,.\rangle$ is the cosine similarity operator, and $B$ represents the mini-batch size. The first term in Eq.~\ref{Eq:OC} ensures intra-identity compactness, while the second term enforces inter-identity separation. Note that, the cosine similarity involves the normalization of fused embeddings, thereby projecting them to a unit hyper-sphere:

\begin{equation}
    \langle \mathbf{l}_{i},\mathbf{l}_{j}\rangle = \frac{\mathbf{l}_{i}.\mathbf{l}_{j}}{\lVert \mathbf{l}_{i} \rVert_{2}. \lVert \mathbf{l}_{j} \rVert_{2}}.
\end{equation}

The joint loss formulation, comprising of $\mathcal{L}_{CE}$ and $\mathcal{L}_{OC}$ as:


\begin{equation}
\label{eq:floss}
    \mathcal{L} = \mathcal{L}_{CE} + \alpha \mathcal{L}_{OC},
\end{equation}

\noindent where $\alpha$ balances the contribution of two terms in $\mathcal{L}$. 

\noindent \textbf{Center Loss.}  It simultaneously learns class centers from features in a mini-batch and penalizes the distance between each class center and corresponding features~\cite{wang2016learning}. Recently, Nawaz et al.~\cite{nawaz2019deep} introduces a Deep Latent Space architecture to extract audio-visual information to bridge the gap between them, leveraging center loss.  The loss is formulated as follow:

\begin{equation}
\label{eq:closs}
\mathcal{L}_{C}=\frac{1}{2} \sum_{i=1}^{b}\left\|\mathbf{I}_{i}-\mathbf{c}_{y_{i}}\right\|_{2}^{2}
\end{equation}

The $\mathbf{c}_{y_{i}}$ denotes $\mathbf{y_{i}}$th class center of features. Wen et al.~\cite{wen2016discriminative} observed that the center loss is very small which may degrade to zeros, thus, it is jointly trained with CE loss as follow:

\begin{equation}
\label{eq:closs_softmax}
    \mathcal{L} = \mathcal{L}_{CE} + \alpha_c \mathcal{L}_{C}
\end{equation}

A scalar value $\alpha_c$ is used for balancing center loss and CE loss.

\noindent \textbf{Git Loss.}  It improves center loss by maximizing the distance between features belonging to different classes (push) while keeping features of the same class compact (pull)~\cite{calefati2018git}. The loss is formulated as follow:

\begin{equation}
\label{eq:git}
\begin{aligned}
\mathcal{L}_{G} &=  \sum_{i,j=1,i\neq j}^b  \frac{1}{1 + \left\Vert \mathbf{I}_{i} - \mathbf{c}_{y_j} \right\Vert^2_2}
\end{aligned}
\end{equation}

Git loss is trained jointly trained with center loss and CE loss as follow:

\begin{equation}
\label{eq:closs_git_softmax}
    \mathcal{L} = \mathcal{L}_{CE} + \alpha_c \mathcal{L}_{C}  + \alpha_g \mathcal{L}_{G}
\end{equation}

Scalar values $\alpha_c$ and $\alpha_g$ are used for balancing center loss, git loss and CE loss.

\section{Experiments}
\label{section:experi}
\noindent \textbf{Training Details and Dataset.} We train our method on Quadro P$5000$ GPU for $50$ epochs using a batch-size of $128$ using Adam optimizer with exponentially decaying learning rate (initialised to \(10^{-5}\)). We extract face and voice embeddings from Facenet~\cite{schroff2015facenet} and Utterance Level Aggregation~\cite{xie2019utterance}. 
We perform experiments on \textit{cross-modal verification} and \textit{cross-modal matching} tasks on the large-scale VoxCeleb$1$ dataset ~\cite{nagrani2017voxceleb}. We follow the same train, validation and test split configurations as used in ~\cite{nagrani2018learnable} to evaluate on \textit{unseen-unheard} identities. We
report results on standard verification metrics i.e. ROC curve (AUC) and equal error rate (EER).

\subsection{Results}
\begin{figure*}

     \centering

     \begin{subfigure}[b]{0.3\textwidth}
         \centering
         \includegraphics[width=\textwidth]{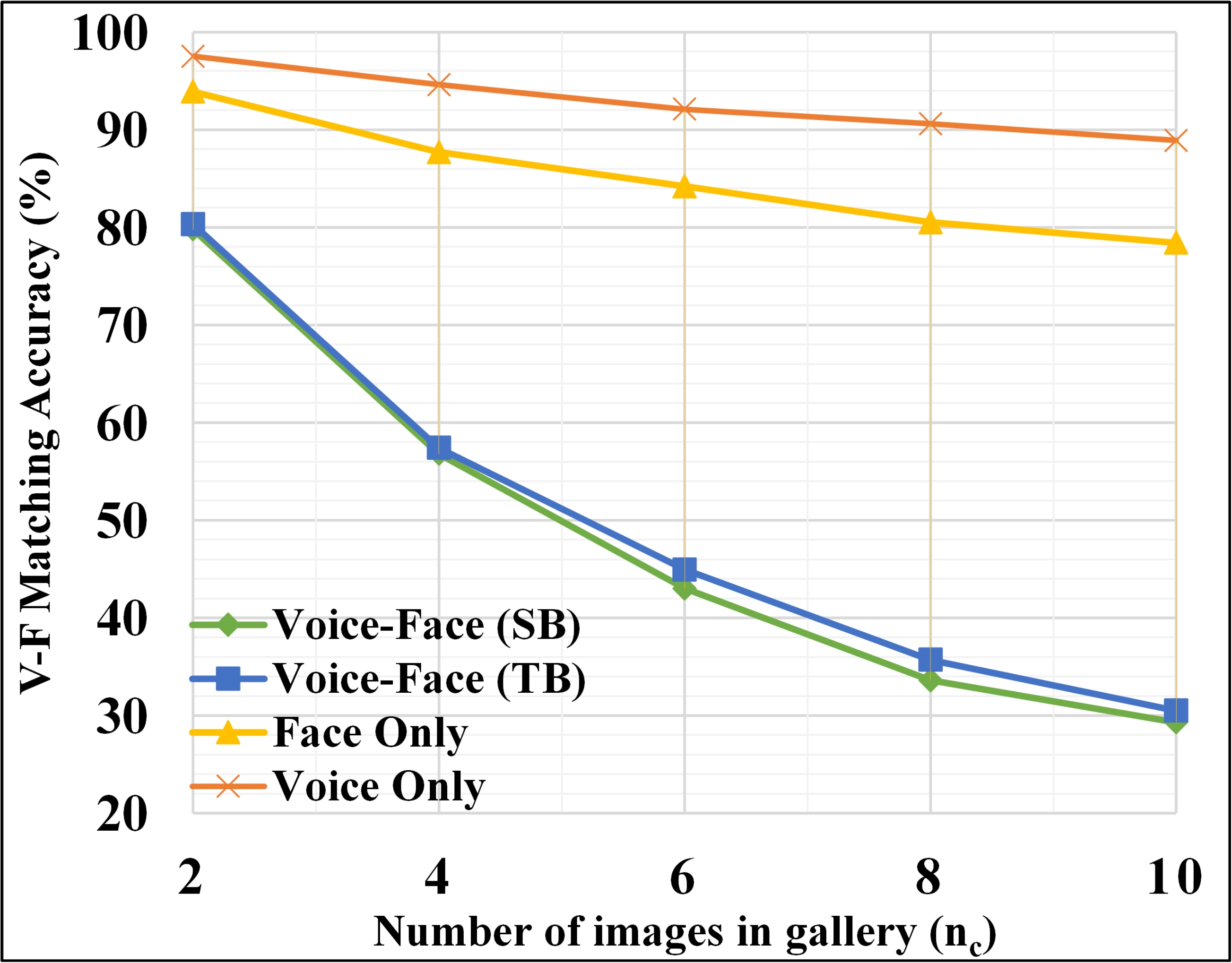}
         \caption{Single and two-branch with FOP}
         \label{fig:y equals x}
     \end{subfigure}
     \hfill
     \begin{subfigure}[b]{0.3\textwidth}
         \centering
         \includegraphics[width=\textwidth]{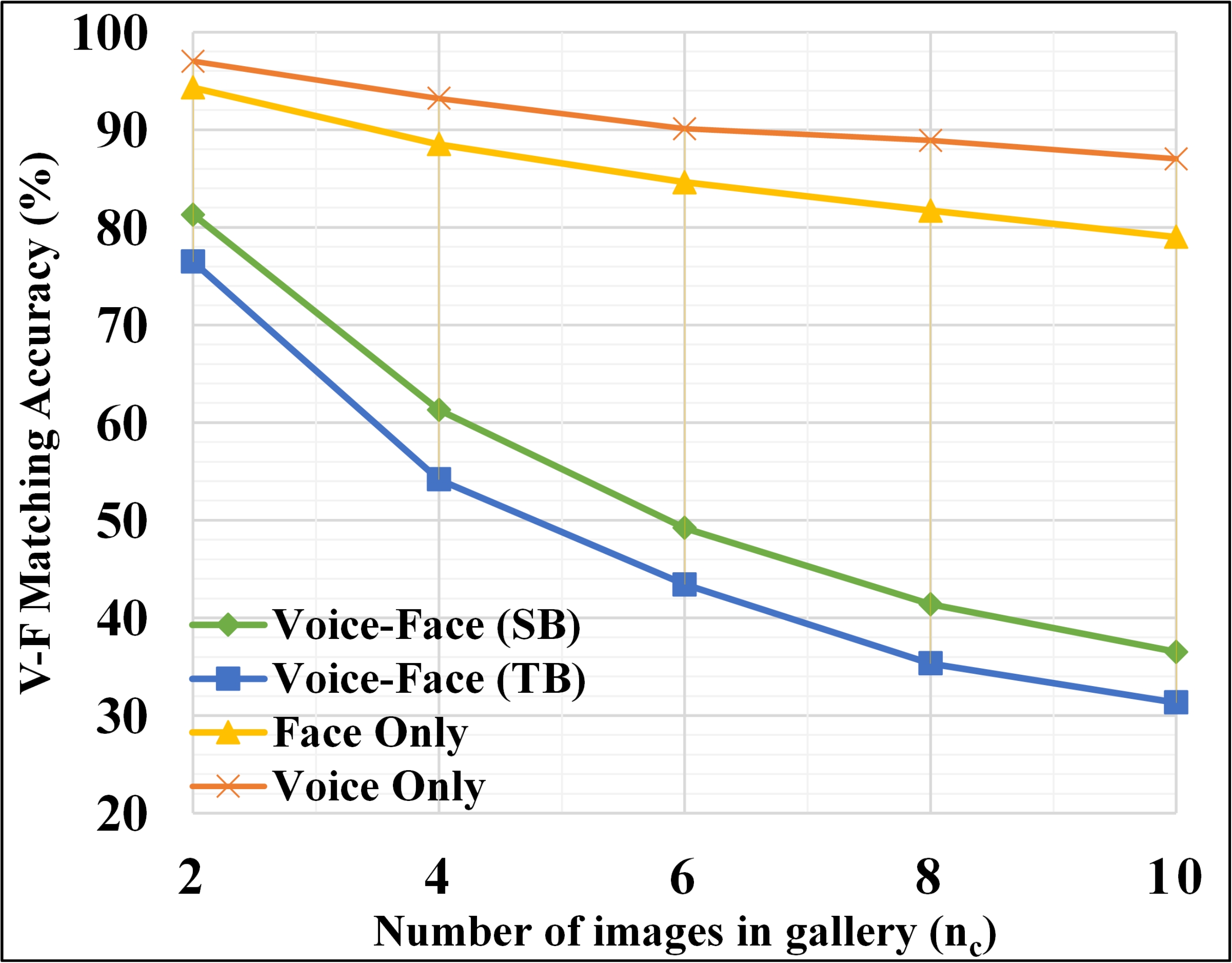}
         \caption{Single and two-branch with Center loss}
         \label{fig:three sin x}
     \end{subfigure}
     \hfill
     \begin{subfigure}[b]{0.3\textwidth}
         \centering
         \includegraphics[width=\textwidth]{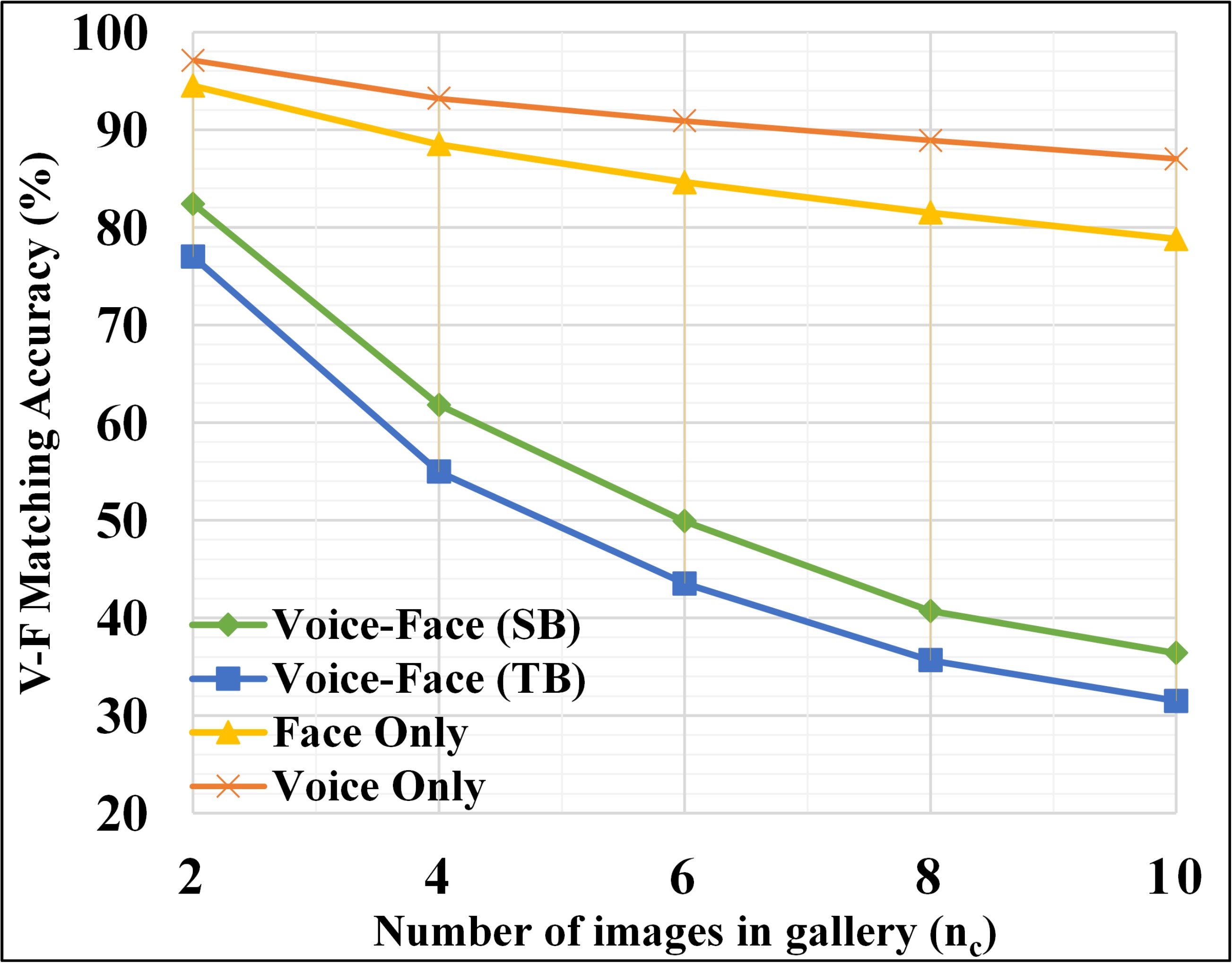}
         \caption{Single and two-branch with Git loss}
         \label{fig:five over x}
     \end{subfigure}
            \caption{ Cross-modal as well as unimodal matching results with varying gallery size for single and two-branch networks using different loss formulations.}
        \label{fig:matching_result}
\end{figure*}

\noindent \textbf{Comparison with two-branch network.} We compare our single-branch with a two-branch network under various loss formulations typically employed in face-voice association, including \textit{fusion and orthogonal projection}~\cite{saeed2022fusion,saeed2022learning}, \textit{center loss}~\cite{wen2016discriminative, nawaz2019deep,saeed2022fusion} and \textit{Git loss} ~\cite{calefati2018git,saeed2022fusion}. 
In addition, we evaluate single-branch network on verification task for face and voice modalities.
Table~\ref{tab:result-base} reveals that our proposed SBNet outperforms two-branch network on all loss formulations.
In addition, we examine effects of Gender (G), Nationality (N), Age (A) and their combination (GNA) separately, which influence both face and voice verification (Table~\ref{tab:results-demographic}). SBNet achieves consistently better performance on  G, N, A, and the combination (GNA) with all loss formulations. Furthermore, we compare our SBNet against two-branch network with various loss functions on a cross-modal matching task, $1:n_c$ with $n_c=2,4,6,8,10$ in Fig.~\ref{fig:matching_result}. We see that it outperforms the counterpart two-branch network with Center and Git loss for all values of $n_c$ whereas comparable performance with FOP.

\begin{table}

\caption{Unimodal and cross-modal verification results on our SBNet and two-branch network using same underlying architecture with various loss formulations.}
\centering
\resizebox{0.99\linewidth}{!}{
\begin{tabular}{|l|c|cc|cc|}
\hline
Train/Test Paradigm  & Loss & \multicolumn{2}{c|}{Single-branch}  & \multicolumn{2}{c|}{Two-branch} \\
\hline
 & & EER & AUC & EER & AUC  \\
 \hline\hline
Face+Voice & \multirow{3}{*}{FOP}  & \textbf{27.5} & 79.7 & 28.0 &\textbf{ 80.0 } \\ 
Voice Only &                       & 8.6  & 97.1 & -    & - \\ 
Face Only  &                       & 14.4 & 93.1 & -    & -  \\ 
\hline
Face+Voice & \multirow{3}{*}{Center}  & \textbf{25.8} & \textbf{81.6} & 31.4 & 75.3 \\ 
Voice Only &                               & 9.6  & 96.7 & -    & - \\ 
Face Only  &                               & 13.1 & 93.9 & -    & -  \\ 
\hline
Face+Voice & \multirow{3}{*}{Git}  & \textbf{25.7} & \textbf{82.4} & 31.3 & 75.5 \\ 
Voice Only &   & 9.6  & 96.7 & - & -\\ 
Face Only  &   & 13.1 & 93.9 & - & -\\ 
 
\hline
\end{tabular}
}

\label{tab:result-base}
\end{table}

\begin{table}[!b]
\caption{Cross-modal biometrics results (AUC) under varying demographics for \textit{unseen-unheard} configurations for our SBNet.}
\centering
\resizebox{0.50\textwidth}{!}{%
\begin{tabular}{|llcccc|lcccc|}
\hline
Demographic  & \multicolumn{5}{c|}{Single-branch}  & \multicolumn{5}{c|}{Two-branch}\\
\hline
             & Rand. & G & N & A & GNA &Rand. & G & N & A & GNA \\
\hline
FOP    & 79.7 & 59.6 & 53.5 & 74.3   & 51.5 & 80.0 & 64.1 & \textbf{53.7}  & 74.3  &  52.1 \\
Center & 81.6 & \textbf{68.6} & 52.0 & 76.1   & 52.7 & 75.3 & 63.6  & 50.8  &  71.3  & \textbf{53.2} \\
Git    & \textbf{82.4} & 68.3 & 52.8 &  \textbf{76.9}  & 52.7 & 75.5 & 63.7  & 50.9  &  71.5  & 52.7  \\
\hline
\end{tabular}}

\label{tab:results-demographic} \vspace{-1.5em}
\end{table}

\begin{figure}[!htp]
\begin{center}
\includegraphics[scale=0.3]{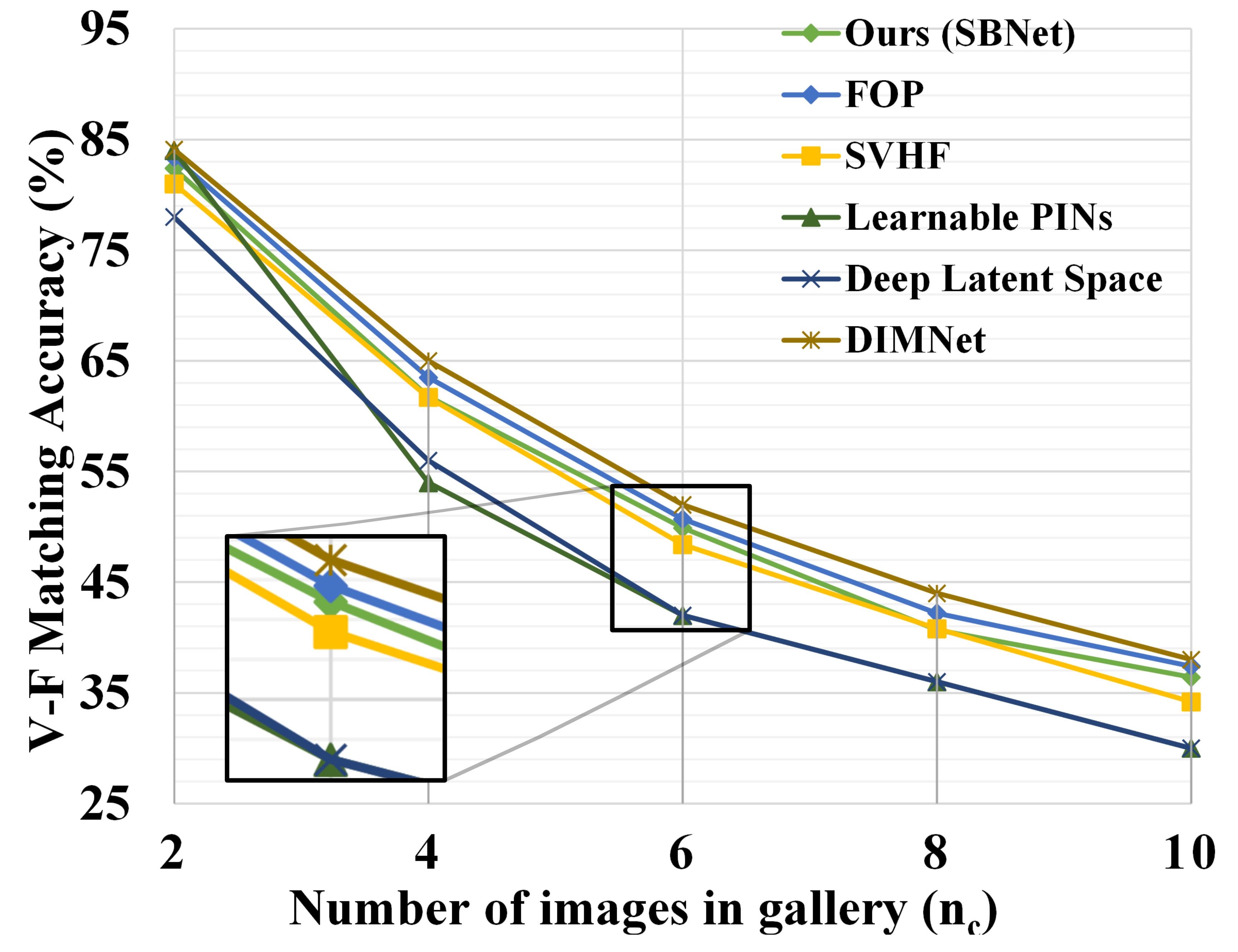}
\end{center} \vspace{-1em}
  \caption{\small Cross-modal matching results of our SBNet and existing SOTA methods with varying gallery size.}
\label{fig:nway-sota}
\end{figure}

\noindent \textbf{Comparison with state-of-the-art.} Table~\ref{tab:sota} compares our method against existing state-of-the-art (SOTA) works (DIMNet~\cite{wen2018disjoint}, Learnable Pins~\cite{nagrani2018learnable}, MAV-Celeb~\cite{nawaz2021cross}, Deep Latent Space~\cite{nawaz2019deep}, Multi-view Approach~\cite{sari2021multi}, Adversarial-Metric Learning~\cite{zheng2021adversarial}, Disentangled Representation Learning~\cite{ning2021disentangled}, and Voice-face Discriminative Network~\cite{tao20b_interspeech}). 
Our SBNet demonstrates comparable performance while utilizing only single branch and similar training paradigm. Among the three losses we compare, Git loss outperforms the others two losses. 
We further compare our SBNet with SOTA works on cross-modal matching (Fig.~\ref{fig:nway-sota}). In particular, we perform the $1:n_c$  matching tasks, where $n_c=2,4,6,8,10$, and report the results. Our SBNet outperforms SVHF, Learnable PINs, and Deep Latent Space by a noticeable margin whereas performs comparably with the rest of the compared methods.

\noindent \textbf{Ablation study and analysis.}
We also analyze the impact of input order during training (Table~\ref{tab:withinepoch-abl}). Several choices are possible including randomly selecting either one of the modalities within an epoch, passing each modality for a portion of epoch (e.g., half epoch face half epoch voice - HeFHeV), or training several epochs on either of the modality before switching to the other (e.g., one epoch voice one epoch face - VFVF).
This also indicates that our single-branch network does not suffer the catastrophic forgetting. To evaluate it further under more challenging scenarios, we experimented by passing one modality for several epochs. Results in Table~\ref{tab:epoch-abl} demonstrate that our approach retains the performance.

\section{Conclusion}
We presented a novel single-branch network to learn both unimodal and multimodal representations. The single-branch uses a series of fully connected layers to extract embeddings of modalities from independent modality specific networks and learns discriminative representations.
Our SBNet out-performed standard two-branch network on all loss formulations on cross-modal verification and matching tasks and performed favourably against the existing SOTA methods on face-voice association tasks.
\begin{table}[!t]
\caption{Cross-modal verification results \textit{unseen-unheard} configurations of our SBNet and existing SOTA methods.}
\centering

\begin{tabular}{|l|cc|}
\hline

Methods     & EER & AUC \\

\hline\hline
DIMNet~\cite{wen2018disjoint}                                       & \textbf{24.9}    & -    \\
Learnable Pins~\cite{nagrani2018learnable}                          & 29.6    & 78.5           \\
MAV-Celeb~\cite{nawaz2021cross}                                     & 29.0    & 78.9           \\
Deep Latent Space~\cite{nawaz2019deep}                          & 29.5    & 78.8            \\
Multi-view Approach~\cite{sari2021multi}                            & 28.0    & -                 \\
Adversarial-Metric Learning~\cite{zheng2021adversarial}             & -       & 80.6              \\
Disentangled Representation Learning~\cite{ning2021disentangled}    & \underline{25.0}  &  \textbf{84.6}  \\
\hline\hline
Single-branch (FOP) - Ours                                          & 27.5     & 79.7      \\
Single-branch (Center) - Ours                                       & 25.8    & 81.6     \\
Single-branch (Git) - Ours                                          & 25.7    & \underline{82.5 }     \\

\hline
\end{tabular}
\label{tab:sota}
\end{table}

\begin{table}[!htp]
\caption{\small Cross-modal verification results when varying input modalities with in each epoch. H(e), V, and F correspond to half epoch, voice, and face respectively. }
\centering
\resizebox{0.65\linewidth}{!}{
\begin{tabular}{|l|c|c|c|c|c|c|}
\hline
Strategy & \multicolumn{2}{|c|}{FOP} & \multicolumn{2}{|c|}{Center loss} & \multicolumn{2}{|c|}{Git loss} \\
\hline
   & EER & AUC & EER & AUC & EER & AUC\\
\hline\hline 
HeFHeV  & 33.8 & 72.1 & 26.0  & 81.6 & 25.8 & 81.9 \\
HeVHeF  & 36.5 & 69.2 & 26.7  & 81.0 & 27.0 & 80.8  \\
Random  & 28.8 & 74.4 & 25.9  & \textbf{82.1} & 25.8 & \textbf{82.5} \\
VFVF    & \textbf{27.5} & 79.7 & 26.0  & 81.6 & 25.8 & 82.4  \\
FVFV    & 27.6 & \textbf{79.9} & \textbf{25.8}  & 81.6 & \textbf{25.7 }& 82.4  \\
\hline
\end{tabular}
}
\hspace{0.08em}

\label{tab:withinepoch-abl}

\end{table}

\begin{SCtable}[][!htp]
\centering
\resizebox{0.3\linewidth}{!}{
\begin{tabular}{|l|c|c|}
\hline
Epoch & \multicolumn{2}{|c|}{FOP} \\
\hline
   & EER & AUC \\
\hline\hline 
1  & 33.2 & 71.1  \\
2  & 32.7 & 71.9  \\
3  & 33.5 & 70.0  \\
5  & 35.2 & 69.5   \\
\hline
\end{tabular}
}
\hspace{0.08em}
\caption{\small Cross-modal verification results when varying epoch. }
\vspace{-3em}
\label{tab:epoch-abl}

\end{SCtable}



%

\bibliographystyle{IEEEbib}
\bibliography{IEEEbib}

\end{document}